\documentclass{article} 
\usepackage{iclr2020_conference,times}

\usepackage{amsmath,amsfonts,bm}

\def\eqref#1{equation~\ref{#1}}

\def\1{\bm{1}}

\DeclareMathAlphabet{\mathsfit}{\encodingdefault}{\sfdefault}{m}{sl}
\SetMathAlphabet{\mathsfit}{bold}{\encodingdefault}{\sfdefault}{bx}{n}

\usepackage{hyperref}
\usepackage{url}
\usepackage{hyperref}       
\usepackage{url}            
\usepackage{amsmath}
\usepackage{booktabs}       
\usepackage{amsfonts}       
\usepackage{nicefrac}       
\usepackage{microtype}      
\usepackage[pdftex]{graphicx}

\title{Classification as Decoder: \\ Trading Flexibility for Control in Neural Dialogue} 

\author{Sam Shleifer\thanks{Work done as research intern at Curai.},\hspace{.1	cm}  Manish Chablani, Namit Katariya, Anitha Kannan, Xavier Amatriain \\
\texttt{sshleifer@gmail.com,  \{manish,namit,anitha, xavier\}@curai.com}}

\iclrfinalcopy 
\begin{document}

\maketitle

\begin{abstract}
Generative seq2seq dialogue systems are trained to predict the next word in dialogues that have already occurred. They can learn from large unlabeled conversation datasets, build a deep understanding of conversational context, and generate a wide variety of responses. This flexibility comes at the cost of control. Undesirable responses in the training data will be reproduced by the model at inference time, and longer generations often don't make sense. Instead of generating responses one word at a time, we train a classifier to choose from a predefined list of full responses. The classifier is trained on (conversation context, response class) pairs, where each response class is a noisily labeled group of interchangeable responses. At inference, we generate the exemplar response associated with the predicted response class. Experts can edit and improve these exemplar responses over time without retraining the classifier or invalidating old training data.

Human evaluation of 775 unseen doctor/patient conversations shows that this tradeoff improves responses. Only 12\% of our discriminative approach's responses are worse than the doctor's response in the same conversational context, compared to 18\% for the generative model. The discriminative model trained without any manual labeling of response classes achieves equal performance to the generative model.
\end{abstract}
\section{Introduction}



Task oriented dialogue systems, exemplified by \citet{multiwoz}, tend to solve narrow tasks like restaurant and hotel reservations and require access to a large knowledge base. After each user utterance,  these systems run multiple modules which parse the user utterance, to try to fill \texttt{(slot, value)} pairs, and pick an action. This setup is too cumbersome for primary care medical conversations, our setting, because (a) building the external knowledge base would require the enumeration of the very large symptom, diagnosis and remedy spaces and (b) each module requires separate training data in large volumes.

The seq2seq group, which we call generative models (GM) \footnote{We use this term to refer to models that generate responses one word at a time, regardless of training objective or decoder sampling method.} require neither labeling nor structured representations of the dialogue state, but manage to learn strong representations of the conversational context with similar content to a knowledge base, according to \citet{petroni2019language}.
They have a key drawback, however: there are no mechanisms to ensure high quality responses. \citet{triggers} show that GPT2 \citep{gpt2} can be attacked with four word sequences to "spew racist output". Many production chatbots check each word in a generated utterance against a blacklist of curse words, but this fails to solve subtler failure modes. Even in a cooperative setting, typos, inaccuracies, and other frequent mistakes in the training data will be reproduced by the model at inference time. \citet{see} find that GM "often repeat or contradict previous statements" and frequently produce generic, boring utterances like "I don't know".

Our discriminative approach attempts to remedy these shortcomings by restricting generations to a manageable set of high quality “exemplar” responses. We ensure that exemplars are all factual, sensible and grammatical by allowing experts to edit them before or after training. For example, if we wanted to switch from recommending users sleep 6-8 hours per night to recommending 7-9 hours, we could simply update the message associated with the output class and the discriminative model would immediately generate the new advice in the same conversational context, without retraining.

Experts can also remove response classes with short, generic exemplars before training to redirect responses towards more productive content. For example the class associated with “that makes sense”, could be removed with the intention of increasing the likelihood of generating “that makes sense. How bad is the pain on a 1-10 scale?”.

We address a key difficulty in this setup -- creating non-overlapping response groups that cover a wide range of situations -- with weak supervision. A pretrained similarity model merges nearly identical responses into clusters, and a human merges the most frequently occurring of these clusters into larger response classes.

To summarize, we propose a system that can generate reasonable responses across multiple domains while restricting generations to a fixed set of high quality responses that are easy to control. We expect our approach to be most useful in task-oriented settings with a wider range of topics, like patient diagnostics and customer service interactions.

The paper is organized as follows: Section \ref{sec:Related} discusses related conversational agents and their methods. Section \ref{sec:Approach} documents our approach, with special attention to the procedure for creating a manageable number of response classes that manage to cover a wide range of conversational contexts. Section \ref{sec: Experiments} explains our main results and the results of experiments which compare the quality of responses suggested by different classification architectures and response class generation procedures.

\section{Related Work}
\label{sec:Related}

%



\textbf{Generative Models:} \citet{hface} won the 2018 NeurIPS PersonaChat competition with "TransferTransfo", a generative transformer approach.
The model starts training with pretrained weights from the GPT2 transformer, then finetunes with the PersonaChat data on a combination of two loss functions: next-utterance classification loss and language modeling (next word prediction) loss. Each task is performed with an associated linear layer that consumes the hidden state of the final self-attention block of the transformer, and both losses are backpropogated into the original transformer weights. Generation is performed with beam search. We compare our architecture to a modified version of this approach in the Section \ref{sec: Experiments}.

\citet{keskarCTRL2019} prepend source URLs to documents during language model pretraining in order to allowing generation conditioned on a source. For example, one could generate in the style of Wikipedia, hopefully preventing expletives and generic utterances. 

\citet{see} directly control repetition, specificity, response relatedness and question asking with a similar conditional training technique. The authors also use weighted decoding, which increases the likelihood of generating tokens that exhibit a certain feature, for example rareness. 

Our approach differs from this stream of research by removing the generative decoding step altogether, and therefore allows direct control of responses and faster inference.


\textbf{Ranking Models}, including \citet{DBLP:journals/corr/abs-1811-01241} and \citet{zhou2018multi}, operate on datasets of \texttt{(context, candidate response, $y$)} triples, where $y$ is a binary label indicating whether the candidate response was the one observed in the conversational data. These models enjoy the same control guarantees as our discriminative approach, since they select responses from a fixed bank, but inference requires both a system to retrieve candidate responses and one forward pass for every candidate response. Our approach chooses between all candidate responses in one forward pass.

\textbf{Discriminative Models:} 

The closest work to ours is \citet{airbnb}'s AirBNB customer service chatbot, which also uses a discriminative approach, but does not attempt to cover the whole response space and differs architecturally. 

Whereas our approach restricts the output space to 187 responses that attempt to cover the whole output space, the AirBNB system chooses from 71 one sentence investigative questions, each representing a cluster of questions, and leaves other response types, like statements and courtesy questions, to a separate model. Although the authors do not report how much of their data these 71 clusters cover, we suspect that it is large, and that their conversations are more homogeneous than ours, since our largest 71 clusters cover only 6\% of our data \footnote{(15\% if we restrict to questions.)} The authors improve class diversity by enforcing that every issue type, a category selected by the user during the chat, is represented by at least one cluster, whereas our approach uses only raw chat logs.

Architecturally, we replace the hierarchical context encoder popularized by \citet{serban} with a larger pretrained language model and show that this approach achieves better performance and much faster training time in Section \ref{sec: Experiments}.

\section{Approach}
\label{sec:Approach}
\subsection{Setup}
We aim to use the last $t$ turns of conversational context to suggest a response for a doctor to send to a patient. Our process involves two stages. First, we create groups of interchangeable doctor utterances, which we call response classes and use as labels for the next step. Then, we train a classifier to predict a response class given the context that preceded it. 

\begin{figure}[!htbp]
\begin{center}
\includegraphics[scale=0.1]{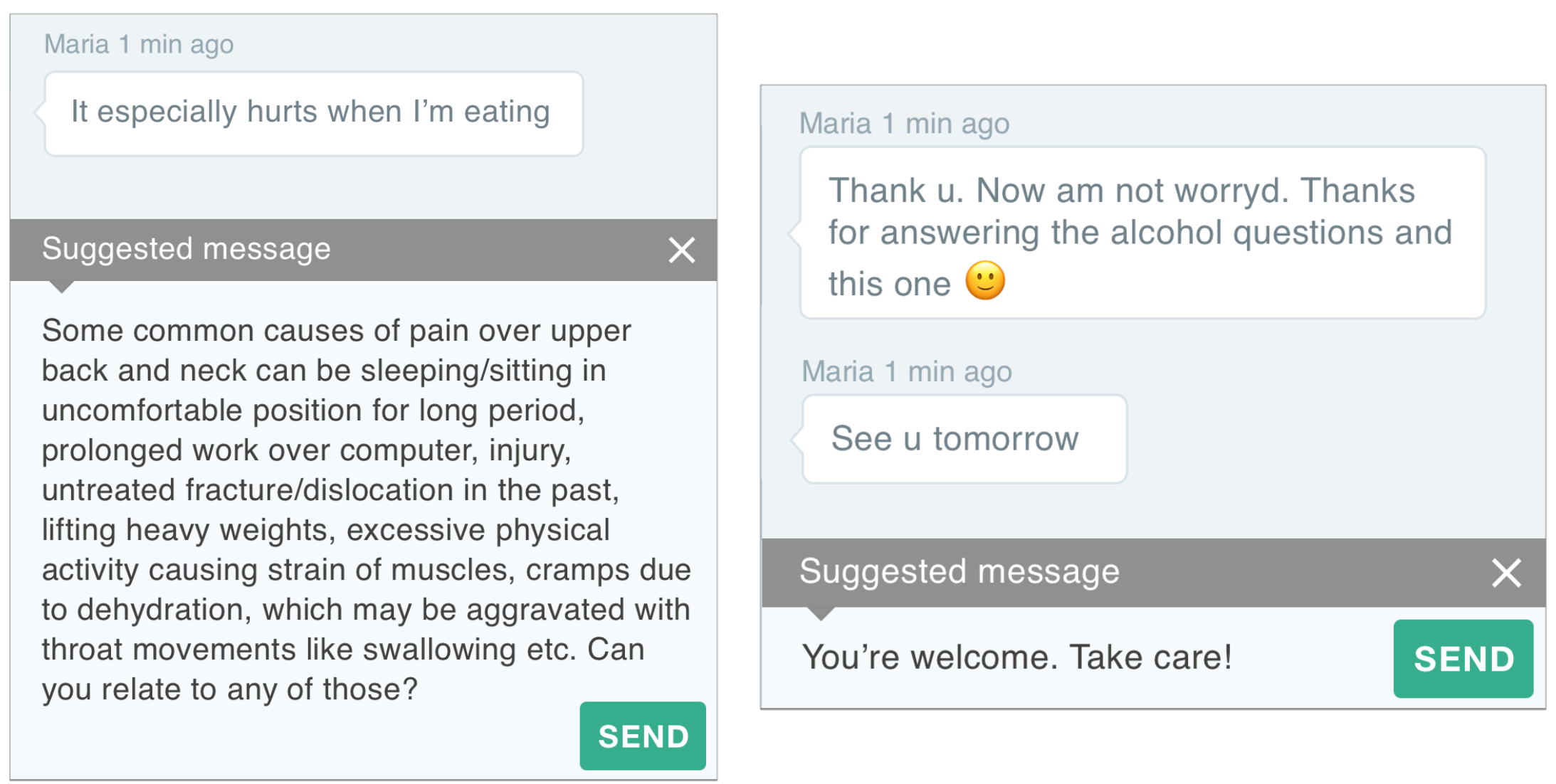}
\caption{A doctor facing suggested response. The model must generate high quality responses in a variety of medical (left) and social (right) contexts.}
\label{screenshot}
\end{center}
\end{figure}

\subsection{Generating Response Classes with Weak Supervision}
\begin{figure}[!htbp]
\begin{center}
\includegraphics[scale=0.4]{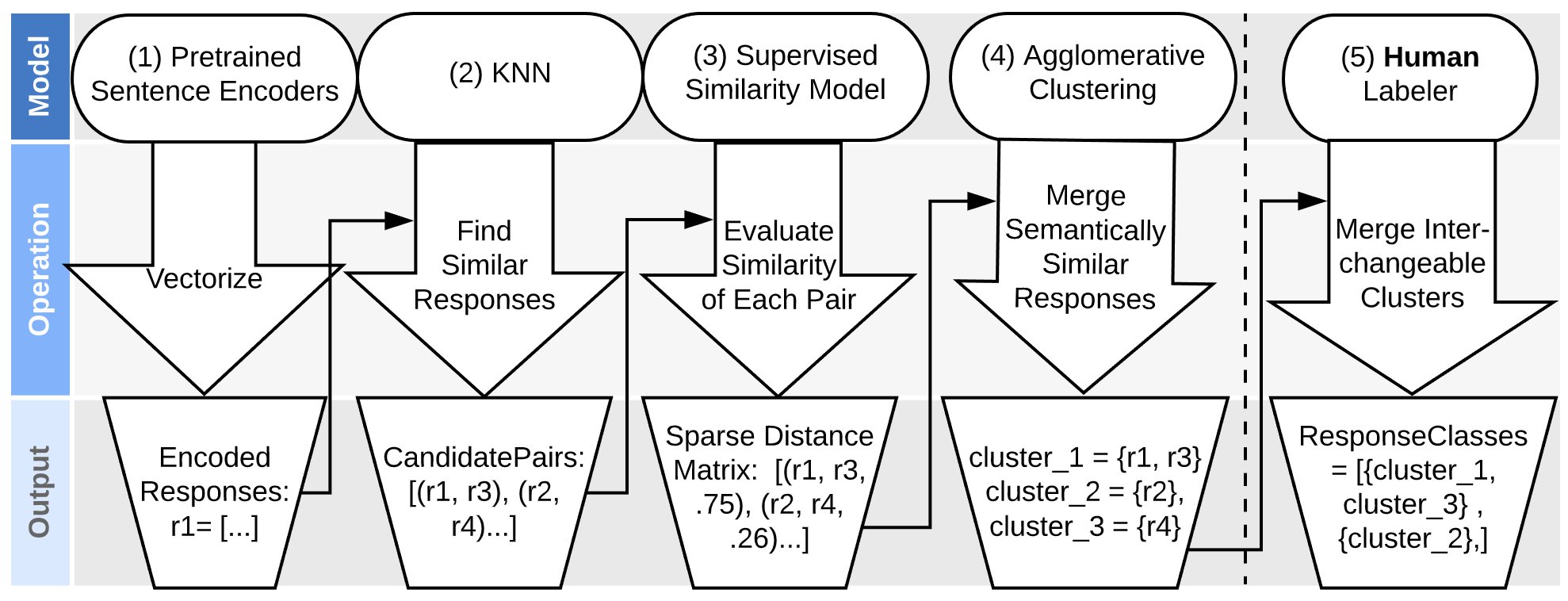}
\caption{Response Class Generation Pipeline}
\label{fig:response_pipe}
\end{center}
\end{figure}
We aim to generate response classes (RCs), where each RC is a group of interchangeable responses observed in the conversation data, with the following characteristics:

\begin{enumerate}
  \item Low overlap between RCs. When interchangeable responses are assigned to different RCs it creates label noise, making the classification task more difficult.
  \item Coverage. If every good response observed in the data is included in an RC, it would increase the likelihood that the classifier can access to a response that matches the conversational context.
\item Sufficient train examples in each RC
\end{enumerate}

We show our five stage procedure in Figure \ref{fig:response_pipe}, and detail the steps below.
First, we lower case all doctor utterances observed in the data, replace patient and doctor identifying information with special tokens, and remove punctuation. We consider only the preprocessed responses $R$ that appear more than once in order to make subsequent steps computationally cheaper.

\textbf{Steps 1 and 2} Since estimating the similarity of every pair of responses is an $O(R^2)$ operation and most pairs are likely to have negligible similarity, we restrict computing similarities to pairs of responses that are within a semantic neighborhood. More specifically, we encode each response $r$ as a vector using $j=4$ different pretrained sentence encoders, and take the 10 nearest neighbors of each responses for each: 
$$
CandidatePairs =  \bigcup_{r \in R, j \in encoders}(r, KNN(enc_{j, r}, enc_{j, R}, 10))
$$
We use InferSent \citep{conneau-EtAl:2017:EMNLP2017}, the finetuned AWD-LSTM language model, the average Glove \citep{pennington2014glove} word vector for the response, and the TFIDF weighted average of the Glove vectors. 

\textbf{Step 3} For each candidate pair, we run a supervised similarity model, BERT \citep{devlin2018bert} pretrained on Quora Question Pairs \citep{Wang_2018}, to predict the probability that each response pair's constituents are semantically similar. We store the dissimilarity of each pair in a sparse distance matrix, with a distance of 1 (the maximum) if two responses were not generated as candidate pairs.
  $$
D_{i, j} =  
\begin{cases}
    (1 - ProbSimilar_{i,j}), & \text{if } (i,j) \in CandidatePairs\\
    1,              & \text{otherwise}
\end{cases}.
$$

\textbf{Step 4:} We run Agglomerative Clustering \citep{scikit-learn} on $D$. 
Two clusters are only merged if \textit{all} the responses in both clusters have $D_{i, j} \leq .25$. The algorithm terminates when no more clusters can be merged.

\textbf{(Optional) Step 5} Manually Merge Clusters into Response Classes
\begin{itemize}
\item Create a dataset where each row represents a cluster, and contains (centroid text, \# occurrences of cluster constituents), sorted by \# occurrences,
where the centroid is the most frequently occurring response in the cluster.
\item For each row in the dataset, the labeler decides whether the cluster centroid text belongs in any existing response class. If so, all constituents of the associated cluster are added to the matching response class. Otherwise, a new response class is created with the constituents of the associated cluster.
\end{itemize}
We merge all responses that have the same impact on the user, and could therefore be used interchangeably. For example, even though "How long have you had the symptoms?" and "When did the symptoms start?" do not mean the same thing, they are both members of the same response class.

If step 5 is skipped, the clusters produced by step 4 are used as response classes.

\subsection{Classification: Conversation Context $\rightarrow$ Response Class}

We train our discriminative response suggestion model to associate a conversational context with the response class observed in the data. \texttt{(Context, ResponseClass)} pairs are only included in the training data if the true response is a member of a response class. We call these included pairs \textit{labeled}.

We follow \citep{ulmfit}'s ULMFit approach with only minor modifications.
Like the original work, we start with an AWD-LSTM language model pretrained on the wiki103 dataset \citep{merity2016pointer}, finetune the language model on our interaction history, and attach a classifier head to predict the response class given the concat pooled representation of final hidden state of the language model.

To accommodate larger batch size than the original work, which we found to help performance, we truncate context sequences to the last 304 tokens before passing them through the language model. This allows us to train with batches of 512 examples.

To encode information about speaker changes, we insert two special tokens: one that indicates the beginning of the patient's turn and one that indicates the beginning of the doctor's turn.

Finally, we add Label Smoothing \citep{DBLP:journals/corr/PereyraTCKH17} with $t=0.1$ to the cross entropy loss function. Label smoothing smooths one-hot encoded classification labels towards $\frac{1}{numClasses}$, and reduces the impact of mislabeled examples on classification training.

\begin{figure}[!htbp]
\begin{center}
\includegraphics[scale=0.2]{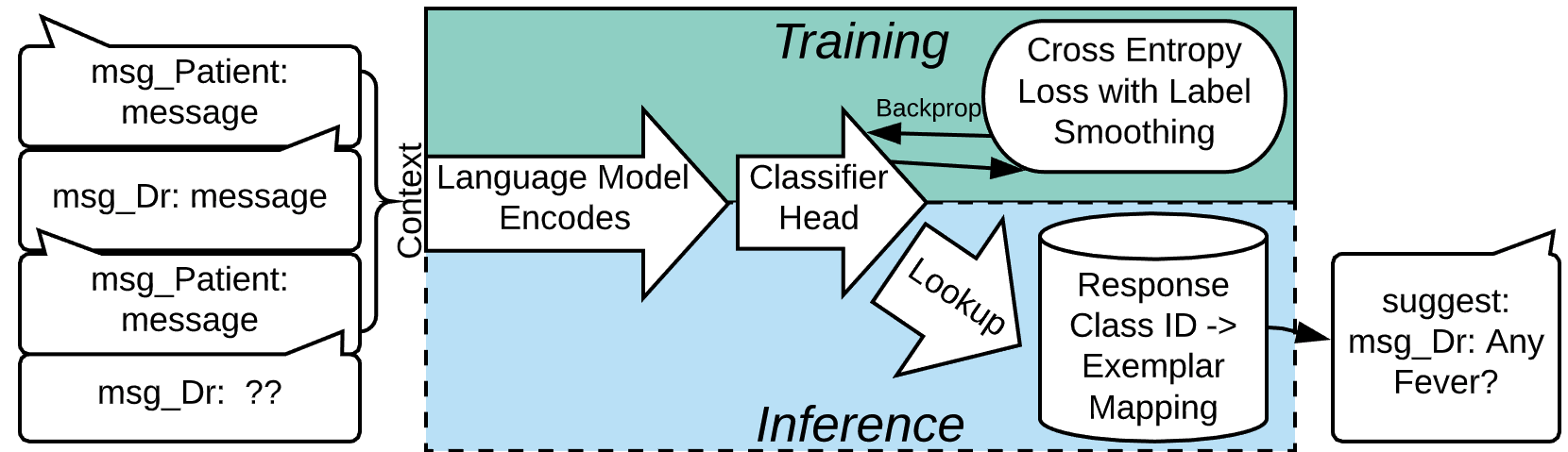}
\caption{Inference and training procedures, starting from a conversational context on the left.}
\label{inference_diag}
\end{center}
\end{figure}

Inference is relatively straightforward. We run the trained classifier and look up the exemplar for the most likely response class, as shown in Figure \ref{inference_diag}.

\section{Experiments}
\label{sec: Experiments}

\subsection{Data}
For language model finetuning, we use 300,000 doctor/patient interactions containing 1.8 million rounds of Doctor/patient exchanges, collected through a web and mobile application for primary care consultations. These exchanges cover a wide variety of medical domains. We use the most recent 100,000 interactions, which contain 700,000 rounds as input to the response class generation procedure, which yields 72,981 labeled pairs for classification training. The number of turns per conversation and length of each turn varies widely, as shown in Table \ref{table1}.

\begin{table}[!htbp]
\centering
\caption{Conversation Level Statistics}
\begin{tabular}{@{}lll@{}}
\toprule
                    & Mean  & Standard Dev.   \\ \midrule
\# Utterances       & 10.8  & 7.85  \\
Words per utterance & 20.4 & 21.8 \\ \bottomrule
\end{tabular}
\label{table1}
\end{table}
\subsection{Clustering Results and Discussion}
\textbf{Clustering Statistics}
Preprocessing and filtering yielded 60,000 frequent responses. Candidate Generation (step 2) yielded 1 million pairs for evaluation. Automated clustering (step 4) yielded 40,000 response clusters with high within-group similarity but many overlapping groups; the largest cluster is 10 distinct responses and 87\% of clusters contain only one response. In the manual step 5, one labeler created 187 groups from the 3,000 most frequently occurring clusters in 3 hours. This leaves roughly 90\% of responses unlabeled.

We hypothesize that our automated clustering procedure leaves many interchangeable clusters unmerged because the sentence encoders were trained to encode a sentence's meaning rather than conversational impact. For example, no encoder produces ("You're welcome. Hoping for the best.", "Take care, my pleasure.") as a candidate pair. 

One advantageous property of our approach is that the manual merging step need not be fully completed, unlike \citet{airbnb}'s. Their KMeans clustering step generates fewer, larger clusters. Curators then go through each cluster and remove heterogeneous constituents, so every response must be considered. This would have taken us 40 hours. \footnote{Another speed vs. label noise tradeoff in our approach: the labeler only considers the new centroid when making a merge decision. Bad merges executed by Agglomerative Clustering are never reviewed, and will result in mislabeled training points. If a centroid is not interchangeable with it's cluster constituents, these errors can silently propagate.}


\subsection{Evaluation}
\label{eval_crit}
\textbf{Expert evaluations for end-to-end comparisons.} To compare discriminative and generative approaches, we construct a test set constructed of unseen \texttt{(conversation, true response)} pairs. Roughly 91\% of test data responses are unlabeled. We call a response unlabeled if it is not an exact duplicate (in it's preprocessed form) of a response one of our 187 response classes.

Given the low correlation of automated metrics such as BLEU score to human judgment of response quality reported in \citet{metrics}, a group of medical doctors evaluated the quality of generated responses on the test data.
For a given conversational context, evaluators compared the doctor response observed in the data to a model's suggested response in a model-blind setting. Evaluators reported whether a model's response is either (a) equivalent to the true response, (b) different but higher quality, (c) different but equal quality, or (d) different but lower quality. For example, "Hoping for the best, take care." and "Take care!!!!" would be marked equivalent. The results are shown in Tables \ref{qresults} and \ref{rset_ablation}. 

\textbf{Accuracy for comparing classifiers.} Tables \ref{archtab} and \ref{ctxtab}, which compare different classifiers on the same dataset, measure accuracy on unseen \textit{labeled} data. 

\subsection{Results}
\begin{table}[!htbp]
\centering
\caption{We find that on 775 test set conversations, the discriminative model compares favorably to the generative model. Evaluators reported whether a model's response could be categorized as options (a) through (d). More details are explained in Section \ref{eval_crit}. Inference time is measured as the average time to generate a response for one suggestion on a V100 GPU with batch size 1. Both models take the last 6 turns of conversation history as input.}
\begin{tabular}{@{}lll@{}}
\toprule
                                & Generative & Discriminative \\ \midrule
a. Equivalent to true response                   & 56\%       & 71\%           \\
b. Different, higher quality & 1\%        & 6\%            \\
c. Different, equal quality  & 25\%        & 11\%           \\ 
d. Different, lower quality & 18\%        & \textbf{12\%}            \\ 
Inference Time   & 800ms         & \textbf{200ms} \\ \bottomrule
\end{tabular}
\label{qresults}
\caption{Architecture comparison: all experiments were completed on a single V100 GPU. $\dag$ follows \citet{hface}, $\S$ follows \citet{serban} and \citet{airbnb}. All QRNN based experiments use random initialization, and 3 layers with hidden size 64. Baselines are discussed in more detail on the next page.}
\begin{tabular}{@{}lllll@{}}
\toprule
Architecture          & 4 turn accuracy & 8 turn accuracy & Encoder Finetune Time & Train Time \\
\midrule
ULMFit              & \textbf{56.7}\%         & \textbf{57.0}\%         & 12h                   & \textbf{40 mins}    \\
QRNN                & 49.3\%         & 49.2\%         & 0                     & 2h         \\
Hierarchical ULMFit & 53.8\%         & 54.9\%         & 12h                   & 18h        \\
Hierarchical QRNN $^\S$  & 47.8\%         & 49.4\%         & 0                     & 6h         \\
Transformer $^\dag$                & 56.6\%         & 56.8\%         & 12h                   & 6h         \\
\bottomrule
\end{tabular}
\label{archtab}
\end{table}

\par \noindent


\textbf{Generative Baseline} For a generative baseline, we use the AWD-LSTM language model \footnote{Pretrained on wiki103, fientuned on our chat logs until convergence (12 hours).} before classification finetuning. For this reason, the context representation used by the classification model head and the language model (generative) head tend to be very similar. We use greedy decoding because beam search was too slow to meet our latency requirements.

\textbf{Hierarchical Baselines} To facilitate comparison with hierarchical encoding (HRED) \footnote{\citet{serban} explain the idea nicely: "The encoder RNN encodes the tokens appearing within the utterance, and the context RNN encodes the temporal structure of the utterances appearing so far in the dialogue, allowing information and gradients to flow over longer time spans."} used by \citet{airbnb}, we tested two different architectures: hierarchical ULMFit (pretrained) and hierarchical QRNN (trained from scratch). In both settings, the higher level context RNN was a randomly initialized QRNN. We found that \textit{flat} ULMFit significantly outperformed its hierarchical counterpart while hierarchical and flat QRNN performed comparably. \citet{serban} who find a 0.18\% accuracy improvement from HRED over a vanilla RNN on their MovieTriples dataset. 

We hypothesize that adding hierarchy to ULMFit decreased performance because of the large variance in the length of turns in our data. Turns vary from 2 to 304 tokens, after truncation, requiring models that consume 3D hierarchical encodings to consume large amounts of padding and smaller batch sizes.\footnote{HRED requires 3D tensors with shape (batch size, max \# of turns, max turn length) so short utterances in conversations that also have long utterances get lots of padding.}
Hierarchical ULMFit on 8 turns could only be trained with batch size 32, while the non-hierarchical ULMFit can fit 512 examples. As mentioned earlier, larger batch size seems to helped ULMFit.

\textbf{Transformer Baseline} To compare with \cite{hface}, we finetune a pretrained double headed transformer on our conversation data, discard the language modeling and multiple choice heads, and attach a one layer classification head that is trained until convergence. As shown in Table \ref{archtab}, this results in similar accuracy to the ULMFit architecture but is much more computationally expensive (10x train time, 20x slower inference).

\par \noindent
 \textbf{How much history is useful?} We find, somewhat counterintuitively, that the ULMFit classifier does not benefit from using more than the last 6 turns of conversation history, as shown in Table \ref{ctxtab}. Similarly, Table \ref{archtab} shows that even for the larger transformer classifier, using the last 8 turns (vs. 4 turns) of history delivers little improvement in classification accuracy.

\begin{table}[!htbp]
\centering
\begin{tabular}{@{}llllllllllll@{}}
\toprule
Max Turns of History   & 1      & 2      & 3      & 4      & 5      & 6      & 7      & All   \\ \midrule
Accuracy & 44.5\% & 53.3\% & 55.3\% & 56.7\% & 56.3\% & \textbf{57.7\%} & 57.4\% & 57.0\%  \\ \bottomrule
\end{tabular}
\caption{One turn is all messages sent consecutively by one conversation participant. Observations are truncated to the most recent $n$ turns.}
\label{ctxtab}
\end{table}

\textbf{Well calibrated probabilities}
\begin{figure}[!htbp]
\begin{center}
\includegraphics[scale=0.2]{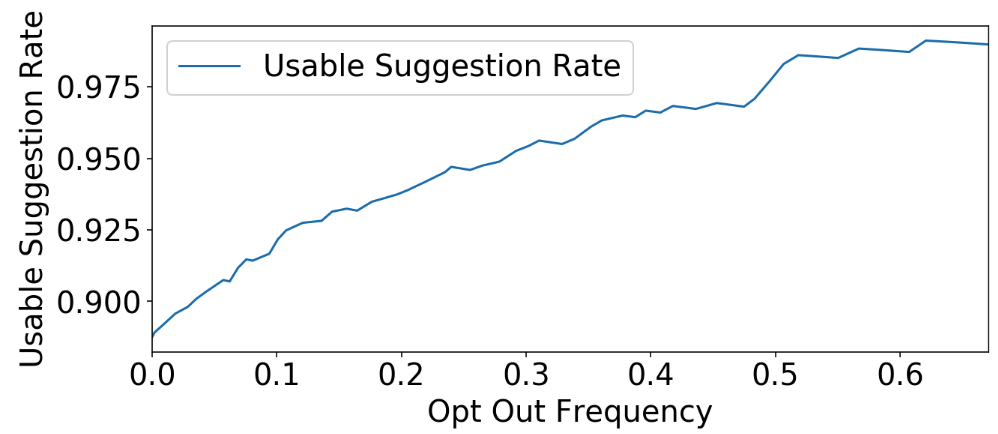}
\caption{The rate of bad suggested responses falls if we "opt-out", and don't suggest any response when the model's predicted probability is low. "Opt Out Frequency" measures how often the model chooses not to suggest a response, while "Usable Suggestion Rate" measures how often the suggested response is not worse than the doctor response observed in the data.}
\label{fig:opting}
\end{center}
\end{figure}

Since the discriminative model must choose from a fixed bank of responses, it will occasionally see context that does not match any of the responses it is trained on. In these circumstances, it should not suggest any reply to the doctor. Figure \ref{fig:opting} shows that if we restrict our evaluations to the 50\% of contexts where the classifier is the most confident (as measured by the maximum predicted probability), the rate of bad suggested responses falls from 11\% to 1.8\%.

\par \noindent
 \textbf{How much manual labeling is needed?}
Table \ref{rset_ablation} shows the response quality produced by discriminative models trained on different datasets. In the 40 class experiment (Row 1), we generate the dataset by following the five step approach described in  Section \ref{sec:Approach}, but stop manually merging clusters after only 20 minutes of work. This results in significantly fewer response classes, less training data, and lower response quality.

A more promising approach takes the other extreme. In the 5,296 class experiment (Row 3), we perform no manual merging of clusters and observe a smaller reduction in response quality, even though classifier accuracy (not shown) falls sharply. In fact, response quality is still slightly higher than the generative model's in this setting.

\begin{table}[!htbp]
\centering
\caption{$^\dag$ Generated with process described in Section \ref{sec:Approach}, including manual merge step. $^\phi$ Same approach with no manual merging step. Bad responses percentage is calculated on 775 test set examples using the the manual evaluation process outlined above. Unique per 100 responses measures how many unique responses are generated per 100 conversation contexts, and is computed on bootstraps from 1000 test set responses. The ground truth data averages 92 unique responses per 100.}
\begin{tabular}{@{}lllll@{}}
\toprule
\# Classes & Weakly Labeled Examples & Bad Responses & Manual Effort & Unique per 100 responses \\ \midrule
40$^ \dag$        & 19,300     & 38\%       &20 Mins     & 17                       \\
187$^ \dag$       & 72,981     & 11\%          & 3 Hours    & 28                       \\
5,296$^ \phi$       & 130,128     & \textbf{17\%}        &0    & 58                       \\ 
Generative & 0 & 18\% & 0 & 86 \\ \bottomrule
\end{tabular}
\label{rset_ablation}
\end{table}

\textbf{Error Analysis}
If evaluators marked a response as "Worse" than the true response, we asked them why. For both models, the primary reason was asking for information that had already been requested or given. This represented roughly 60\% of errors for both the generative and discriminative (187 class model). Roughly 15\% of GM errors were about nonsensical responses (roughly 15\% of  errors), including "It is not advisable to take Sudafed with Sudafed". The discriminative model has no nonsensical responses, but generates "You're welcome" in nearly every sign-off, even if the patient never said "thank you", a mistake the generative model never makes. 


\section{Conclusion}
In this work, we propose a classification model that leverages advances in pretraining techniques to generate useful responses in a wide variety of contexts while restricting generations to a fixed, easy to update set of high quality responses. This allows full control of generations, but  also helps the average suggested response quality, compared to a generative model with a nearly identical backbone.

The key error source for both models, asking the same question twice could be more easily fixed for the discriminative model, by using the second most likely class if the most likely class has already been generated in a conversation. For the generative model, this could be addressed with conditional training, following \citet{see}.

The key difficulty in this approach, and opportunity for future work, is the creation of response classes. We might be able to obviate the need for a manual merging step, for example, by allowing creation (through merging) of clusters with pairs of constituents that were not generated as candidates or by training a separate classifier to predict which response class an unseen response belongs to.

Finally, we intend to test whether the control for flexibility tradeoff provides similar quality improvements in other conversational domains.

\bibliography{main_iclr}
\bibliographystyle{iclr2020_conference}


\end{document}